\documentclass[10pt,twocolumn,letterpaper]{article}

\usepackage[pagenumbers]{cvpr} % To force page numbers, e.g. for an arXiv version
\usepackage{multirow}
\usepackage{multicol}
\definecolor{cvprblue}{rgb}{0.21,0.49,0.74}
\usepackage[pagebackref,breaklinks,colorlinks,allcolors=cvprblue]{hyperref}

\def\confName{CVPR}
\def\confYear{2025}

\def\ourapproach{\textsc{xRIR}\xspace}
\def\ourdataset{\textsc{AcousticRooms}\xspace} 

\title{Hearing Anywhere in Any Environment}

\author{
Xiulong Liu $^{1}$\hspace{1pt}\thanks{Work done during the internship at Meta.} \quad
Anurag Kumar$^{2}$ \quad
Paul Calamia$^{2}$ \quad
Sebastià V. Amengual$^{2}$ \quad
Calvin Murdock$^{2}$ \\
Ishwarya Ananthabhotla$^{2}$ \quad
Philip Robinson$^{2}$ \quad
Eli Shlizerman$^{1}$ \quad
Vamsi Krishna Ithapu$^{2}$ \quad
Ruohan Gao$^{3}$\hspace{1pt}\\[0.5em]
$^{1}$University of Washington \quad
$^{2}$Meta \quad
$^{3}$University of Maryland, College Park
}

\begin{document}
\maketitle

\begin{abstract}
In mixed reality applications, a realistic acoustic experience in spatial environments is as crucial as the visual experience for achieving true immersion. Despite recent advances in neural approaches for Room Impulse Response (RIR) estimation, most existing methods are limited to the single environment on which they are trained, lacking the ability to generalize to new rooms with different geometries and surface materials. We aim to develop a unified model capable of reconstructing the spatial acoustic experience of any environment with minimum additional measurements. To this end, we present \ourapproach, a framework for cross-room RIR prediction. The core of our generalizable approach lies in combining a geometric feature extractor, which captures spatial context from panorama depth images, with a RIR encoder that extracts detailed acoustic features from only a few reference RIR samples. To evaluate our method, we introduce \ourdataset, a new dataset featuring high-fidelity simulation of over 300,000 RIRs from 260 rooms. Experiments show that our method strongly outperforms a series of baselines. Furthermore, we successfully perform sim-to-real transfer by evaluating our model on four real-world environments, demonstrating the generalizability of our approach and the realism of our dataset.
\end{abstract}

\begin{figure}[tp]
    \centering
    \includegraphics[width=\linewidth]{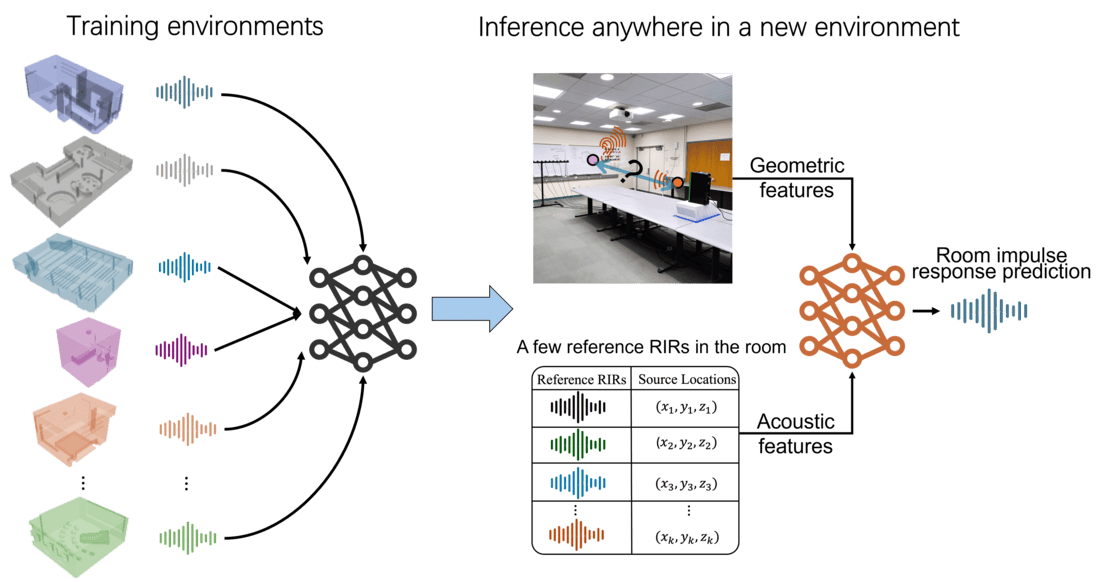}
    \vspace{-0.25in}
    \caption{Our \ourapproach framework can predict accurate room impulse responses (RIR) of any new environment, by integrating the geometric prior learned from a large simulated RIR dataset of diverse training environments and the nuanced acoustic profile extracted from a few reference RIR measurements in the new environment. 
    }
    \vspace{-4mm}
    \label{fig:teaser}
\end{figure}

\section{Introduction}
\label{sec:intro}
Each environment echoes its own story, creating a distinct auditory experience. Imagine walking through a museum where each room's unique acoustic character brings exhibits to life, immersing you in stories through tailored soundscapes. To recreate such realistic auditory experiences across different spaces, models must seamlessly and easily adapt from one environment to another. This \emph{cross-room} adaptability is essential for applications like virtual reality and immersive media, where authentic acoustics transform and enhance how we experience sound anywhere in any environment of interest.

To model how sound interacts with an environment, a room impulse responses (RIR) is often measured to capture how a perfect impulse emitted from the source location reflects, is absorbed, diffuses, and gets received at the microphone location, all according to the room's unique geometry and surface materials. Traditionally, to fully capture the immersive acoustic field in an environment, hundreds of RIR measurements are gathered by positioning speakers and microphones densely throughout the space~\cite{richard2022deep, ratnarajah21_interspeech, luo2022learning, su2022inras}. However, this process is labor-intensive and costly, especially when scaling across diverse real-world environments. 

Recent deep learning approaches~\cite{luo2022learning,su2022inras,liang2023neural,chen2024real} leverage implicit neural networks to ``compress'' these dense RIR measurements into a single model that can be queried for RIR at any source-listener pair in a specific room. However, training these models still requires a large amount of densely sampled RIRs for each room, as they are designed to overfit to a \emph{single room}'s specific geometry and material properties. When applied to a new environment, these models must be re-trained with a similarly dense dataset, limiting their practicality for scalable cross-room applications.

Our goal is to develop a model that can handle variations in room geometry and surface material properties, which are key factors in shaping each room's unique acoustic profile. Achieving this goal presents several challenges. First, the model should utilize an easily obtainable and standard visual representation of the environment to extract geometric properties from any room. Second, given the diverse and nuanced acoustic properties of different surface materials, we need a way to quickly capture essential cues about a room's detailed acoustic characteristics (\eg, energy decay and reverberation patterns). Third, to ensure generalization, we need a large-scale and high-fidelity RIR dataset that encompasses a wide range of room environments with varied acoustic materials and geometries, enabling the pre-training of a cross-room feature extractor.

To address these challenges, we introduce \ourapproach, a generalizable model for cross-room RIR prediction, along with \ourdataset, a large dataset comprising over 300,000 realistic RIRs simulated from 260 rooms, specifically curated for this task. Our model features three key components: i) a Geometric Feature Extractor, which utilizes a vision transformer to process panorama depth images from the receiver's perspective, capturing the spatial relationships between the source and receiver positions within the room; ii) a Reference RIR Encoder, which extracts spatio-temporal features from a few reference RIRs, capturing the unique energy decay and reverberation characteristics associated with room materials; and iii) a Fusion and Weighting Module, which predicts the target RIR through a weighted combination of the reference RIRs. By integrating complementary geometric and acoustic features, our model effectively approximates both structural and material properties of any room, enabling precise RIR predictions not only at new locations in the training environments, but also in any new environment of interest.

We evaluate our model's performance in both seen and unseen environments from \ourdataset, demonstrating its capability to predict RIRs at new locations within known rooms and also effectively generalize to entirely new environments. Our method consistently achieves state-of-the-art results across these scenarios, outperforming several strong baselines and prior methods. In addition, to assess the model's real-world applicability, we successfully perform sim-to-real transfer by deploying our model on a dataset comprising four real-world environments. This demonstrates the effectiveness of our framework and also the realism of RIR simulations in our dataset.

In sum, our main contributions are as follows:
\begin{itemize} [align=right,itemindent=0em,labelsep=2pt,labelwidth=1em,leftmargin=*,itemsep=0em] 

\item We propose \ourapproach, a cross-room generalizable framework that predicts accurate RIRs for any seen and unseen environment, strongly outperforming prior methods.

\item We introduce \ourdataset, a large-scale dataset tailored for this task, comprising over 300,000 high-fidelity RIRs simulated from 260 diverse rooms.

\item Apart from superior performance on simulated rooms, we also successfully deploy our model in four real-world environments, showcasing effective sim-to-real transfer.

\end{itemize}

\section{Related Work}
\label{sec:related_works}

\noindent\textbf{Learning-Based RIR Prediction.}
Early machine learning methods for room impulse response (RIR) prediction, such as Image2Reverb~\cite{singh2021image2reverb} and Fast-RIR~\cite{ratnarajah2022fast}, utilize a generative approach conditioned on semantic information like RGB images of the environment, source and listener locations, and T60 values. While these approaches produce plausible RIRs aligned with scene semantics and basic acoustic constraints, they struggle to reproduce accurate RIRs at arbitrary locations within the target scene. Recent advances on implicit neural representations~\cite{sitzmann2020implicit,mildenhall2021nerf} have inspired a series of works that approximate a function mapping spatial coordinates of source and listener locations to RIRs~\cite{su2022inras,luo2022learning,ratnarajah2022mesh2ir,richard2022deep}. Some methods~\cite{liang2023av,ratnarajah2024av,bhosale2024av,chen2024real,ahn2023novel,chen2024avcloud,lan2024acoustic} also condition on room geometry and material properties of the visual environment. By explicitly modeling the 3D scene, these methods can render precise RIRs at novel locations within the same environment they are trained on. However, they lack the ability to generalize to new environments, which is the focus of our work.

Closest to our work are Diff-RIR~\cite{wang2024hearing} and Few-Shot RIR~\cite{majumder2022few}, both of which also address cross-room RIR generalization. Diff-RIR learns material coefficients through a differentiable rendering framework based on planar room geometry and a few RIR measurements, enabling RIR rendering at any location using the image source method. However, it requires training a separate model for each room, making it computationally intensive and less scalable for large or complex environments. Few-Shot RIR leverages a limited number of RIR measurements and RGB-D observations to predict RIRs at new locations by integrating features from pose, RGB-D data, and binaural echoes. However, it does not effectively utilize geometric information as our method and is only trained and tested in simulation. We compare our approach with both methods in experiments.
\vspace{0.05in}

\noindent\textbf{Room Acoustics Simulation.}
Room acoustic simulation is crucial for applications in AR/VR~\cite{lentz2007virtual}, architectural design~\cite{peters2015integrating}, and far-field speech recognition~\cite{haeb2020far,ratnarajah2020ir}, serving as a bridge between simulated and real-world acoustics. Approaches can be generally categorized as wave-based, geometric-based, or hybrid. Wave-based methods~\cite{deckers2014wave} accurately capture low-frequency phenomena but are computationally intensive. Geometric-based methods, such as~\cite{scheibler2018pyroomacoustics, chen2020soundspaces, chen2022soundspaces, 10.1145/3450626.3459751}, efficiently trace sound rays but lack accuracy at low frequencies. Hybrid methods~\cite{tang2022gwa} combine both approaches to balance accuracy and efficiency. We utilize the hybrid approach~\cite{osti_4491151} from the Treble simulation platform to simulate high-fidelity RIRs for our cross-room RIR prediction task, optimizing both accuracy and computational demands.
\vspace{0.05in}

\noindent\textbf{Audio-Visual Learning of Room Acoustics}. Both vision and audio provide significant spatial information that reveals room properties. Prior methods have combined both modalities for a series of audio-visual learning tasks related to room acoustics, including audio spatialization using visual spatial cues from the environment~\cite{garg2021geometry,morgado2018self,garg2023visually,gao2019visual-sound,li2018scene}, audio-visual navigation in environments with varying room acoustic properties~\cite{chen2021waypoints,chen2020soundspaces,gao2023sonicverse,gan2020look,Liu_Paul_Chatterjee_Cherian_2024}, learning image features, scene structures, or human locations from echoes~\cite{christensen2020batvision,gao2020visualechoes}, ambient sound~\cite{chen2021structure}, or music~\cite{wang2024soundcam} in the room, and using RGB images or videos of a target environment to guide sound transfer that aligns with the space's acoustics~\cite{chen2022visual,li2024self,chen2023novel, somayazulu2024self,chowdhury2023adverb}. Our work also integrates both visual and audio information, but we address a different challenging task that aims to infer accurate room acoustics in any new environment.

\section{Our Approach}
\label{sec:methods}
\subsection{Problem Formulation}
\label{sec:task}

We tackle the \emph{cross-room} room impulse response (RIR) prediction task, which aims to predict single-channel (omnidirectional) RIRs for any source-receiver pair across diverse room environments, including those \emph{unseen} during training. We aim to develop a generalizable model that can accurately predict RIRs in any environment without labor-intensive data collections or training a separate model for each room.  Our model (detailed in Sec.~\ref{sec:model}) achieves this by utilizing only minimal additional measurements from the new room, such as only a few panorama depth images and reference RIR measurements, to quickly adapt to new acoustic environments with minimal effort, thereby facilitating generalization to previously unseen environments. Next, we formally define the cross-room RIR prediction task by outlining its data, inputs, and the modeling objective.

 \textbf{Data.} Let \( R = \{ R_1, R_2, \dots, R_M \} \) represent a dataset of \( M \) rooms, split into a training set, \( R_{\text{train}} \subset R \), and a test set, \( R_{\text{test}} = R \setminus R_{\text{train}} \). Each room \( R_m \) includes a set of receiver locations, denoted \( L_m = \{ P_{r}^{(m,1)}, P_{r}^{(m,2)}, \dots, P_{r}^{(m,N_m)} \} \), where \( N_m \) is the number of receivers in the room. For each receiver \( P_{r}^{(m,i)} \) in room \( R_m \), RIRs are measured at various source locations \( P_{s}^{(m,i,j)} \), resulting in measurements \( A_{m,i,j} \) of the source-receiver pair  \( P_{s}^{(m,i,j)} \) and \( P_{r}^{(m,i)}\) .

 \textbf{Inputs.} To capture the necessary observation conditions for predicting a target RIR \( A_t \), we define an observation tuple \( \mathbf{O} = (P_s, P_r, G_r) \), where \( P_s \) is the target source location, \( P_r \) is the receiver location, and \( G_r \) represents the local geometry near the receiver location \( P_r \), e.g., room boundary points or depth maps around the receiver. Additionally, we introduce a set of \( K \) reference RIRs measured at the target receiver location \( P_r \) from various reference source locations \( \mathbf{P}_{\text{ref,s}} = \{ P_{\text{ref,s}}^{(1)}, P_{\text{ref,s}}^{(2)}, \dots, P_{\text{ref,s}}^{(K)} \} \). These references, denoted as \( \mathbf{A}_{\text{ref}} = \{ A_{\text{ref}}^{(1)}, A_{\text{ref}}^{(2)}, \dots, A_{\text{ref}}^{(K)} \} \), are crucial for capturing essential acoustic characteristics that encode nuanced information about room materials. Note that in the above formulation, while we fix the receiver location and set reference RIRs at different source locations, exchanging the receiver and source in the input yields \textit{an equivalent alternative formulation}.

\textbf{Modeling Objective.} The objective of cross-room RIR prediction is to train a model \( F \) that predicts the target RIR \( \hat{A}_t \) using the observation tuple \( \mathbf{O} \) along with the reference RIRs \( \mathbf{A}_{\text{ref}} \) and their respective source locations \( \mathbf{P}_{\text{ref,s}} \):
$\hat{A}_t = F(\mathbf{O}, \mathbf{A}_{\text{ref}}, \mathbf{P}_{\text{ref,s}})$.
In this formulation, \( \mathbf{O} \) provides the geometric and positional context, while \( \mathbf{A}_{\text{ref}} \) and \( \mathbf{P}_{\text{ref,s}} \) give sparse acoustic observations that help bridge the lack of explicit material properties by capturing key room acoustics characteristics.

Unlike the \emph{single-room} RIR prediction task~\cite{luo2022learning,su2022inras}, which assumes consistent geometry and material properties and fits a separate model for each scene, our cross-room formulation aims to train a single model that generalizes across multiple scenes with diverse room geometries and materials, while with the extra condition of only a few RIR measurements. Models that are designed for single-room RIR prediction task ~\cite{su2022inras, luo2022learning} must be re-trained with dense data when applied to a new environment. While our method uses one unified model to predict accurate RIRs across different rooms, seen or unseen.
Our formulation can also be easily adapted to the single-room RIR prediction setting by fitting dense measurements in the room as in prior work. Please see supp. for results on single-room experiments.

\subsection{The \textbf{\ourapproach} Model}\label{sec:model}

\begin{figure*}[tp]
    \centering
    \includegraphics[width=\linewidth]{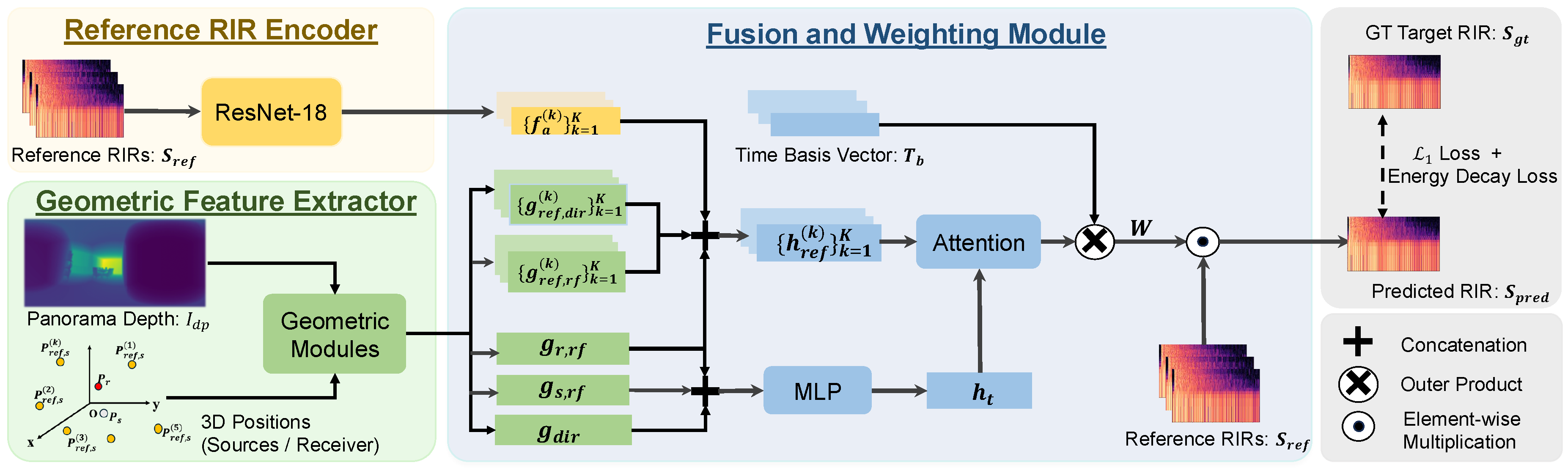}
    \vspace{-0.2in}
    \caption{\textbf{System Overview of Our \ourapproach Model for Cross-Room RIR Prediction.} The model architecture consists of three main components: i) a \emph{Geometric Feature Extractor}, which captures spatial relationships among the source, receiver, and room geometry; ii) a \emph{Reference RIR Encoder}, which extracts spatiotemporal features from reference RIRs; and iii) a \emph{Fusion and Weighting Module}, which integrates these spatial and acoustic features to predict the target RIR.}
    \vspace{-2mm}
    \label{fig:system_overview}
\end{figure*}

To solve the cross-room RIR prediction task, we propose a new architecture, \textit{\ourapproach}, which processes not only geometry and positional features of source and receiver, but also leverages the reference RIRs to accurately predict the target RIR. As illustrated in Fig.~\ref{fig:system_overview}, \ourapproach consists of three main components: i) a \emph{Geometric Feature Extractor} (Sec.~\ref{geometric_feature_extractor}), which encodes the spatial relationships among the source, receiver, and room surface geometry, capturing important geometric features that shape acoustic behavior; ii) a \emph{Reference RIR Encoder} (Sec.~\ref{reference_RIR_encoder}), which processes the spatiotemporal characteristics of the reference RIRs to extract features that represent their acoustic properties within the room.; and iii) a  \emph{Fusion and Weighting Module} (Sec.~\ref{fusion_and_weighting}), which integrates the spatial features from the Geometric Feature Extractor with the reference RIR features from the Reference RIR Encoder, generating a set of weights to combine reference RIRs as the predicted target RIR.

\subsubsection{Geometric Feature Extractor}\label{geometric_feature_extractor}

\begin{figure}[tp]
    \centering
    \includegraphics[width=\linewidth]{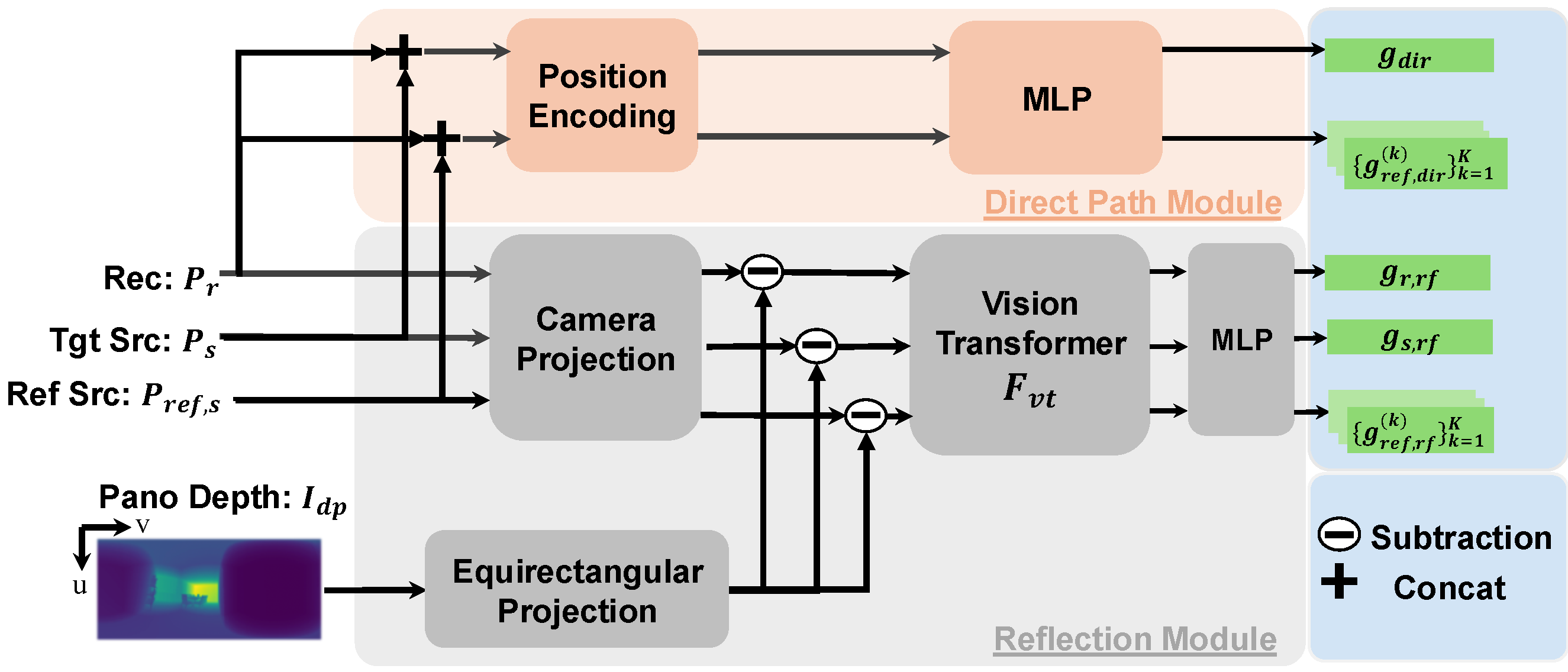}
    \vspace{-0.2in}
    \caption{\textbf{Illustration of the Geometric Feature Extractor.} Rec: Receiver, Tgt Src: Target Source, Ref Src: Reference Source.}
    \vspace{-3mm}
    \label{fig:geometric_feature_extraction}
\end{figure}

The Geometric Feature Extractor module captures spatial relationships among the source, receiver, and room geometries, which is important for accurate acoustic modeling. It consists of two geometric sub-modules: the \emph{Direct Path Module}, and the \emph{Reflection Module}. These two modules emulate the process of sound propagation. The Direct Path Module extracts the feature of the direct path between source and receiver, while the Reflection Module models the sound propagation path through the reflections from the room boundaries. A detailed overview of the Geometric Feature Extractor is illustrated in Fig.~\ref{fig:geometric_feature_extraction}.

\textbf{Direct Path Module.} To capture the direct path between each source and the receiver, we concatenate their 3D coordinates. For the target source, we define \( P_{\text{dir}} = (P_s, P_r) \), where \( P_s \) and \( P_r \) are the coordinates of the source and the receiver, respectively. For each reference source \( P^{(k)}_{\text{ref,dir}} \), we define \( P^{(k)}_{\text{ref,dir}} = (P^{(k)}_{\text{ref},s}, P_r) \). \( P_{\text{dir}} \) and each \( P^{(k)}_{\text{ref,dir}} \) encode locations of every source-receiver pair, thereby encoding the direct path information. To extract their features, we apply sinusoidal positional encoding~\cite{NIPS2017_3f5ee243} followed by a multi-layer perceptron (MLP) to project them into high-dimensional vectors, resulting in \( g_{\text{dir}} \in \mathbb{R}^{1 \times C_d} \) for the target source-receiver pair and \( g^{(k)}_{\text{ref,dir}} \in \mathbb{R}^{1 \times C_d} \) for each reference source-receiver pair.

\textbf{Reflection Module.} Inspired by INRAS~\cite{su2022inras}, we also model the reflection paths between source and receiver via the room boundary. But differently, instead of using a fixed set of bounce points per room, we propose to use a panorama depth map at the receiver's location as a proxy for local room geometry to unify the representation across different rooms. 

Given the panorama depth map $I_{dp} \in \mathbb{R}^{H \times W \times 3}$ centered at receiver's viewpoint, we first project $I_{dp}$ to a 3D coordinate map \( I_{\text{coord}} = F_{\text{rect}}(I_{dp}) \) via an equi-rectangular projection transformation $F_{\text{rect}}$. Each pixel in \( I_{\text{coord}}  \) represents the 3D coordinate of a visible boundary point in the room from the receiver's view. Each source in the room has chance to reflect through these points until finally reaching the receiver. To model such interactions, we create a set of reflections-based maps by subtracting the 3D coordinate map from the source and receiver positions. 

To perform the subtraction, it is necessary to unify the coordinates between the 3D coordinate map and the source / receiver positions. We achieve this by projecting the world coordinates of the sources and the receiver into camera coordinates at the receiver's position, resulting in the same coordinate system as the 3D coordinate map. We obtain the target source position as \( P_{\text{rel}, s} = R (P_s - P_r) \) and each reference source location as \( P_{\text{rel, ref}}^{(k)} = R (P_{\text{ref},s}^{(k)} - P_r) \), where $R$ is the world-to-camera transformation matrix. Then we create reflection-based maps \( I_{\text{rf}, s} \) for the target source, \( I_{\text{rf, ref}}^{(k)} \) for reference sources, as well as \( I_{r,\text{rf}} \) for the receiver by performing subtractions: \( I_{s,\text{rf}} = P_{\text{rel}, s} - I_{\text{coord}} \), \( I_{\text{ref,rf}}^{(k)} = P_{\text{rel, ref}}^{(k)} - I_{\text{coord}} \) and \( I_{r,\text{rf}} = I_{\text{coord}} - \mathbf{0} \), where $\mathbf{0}$ means the origin, where the receiver is located. 

These reflection-based maps encode the dense interaction between room geometry and the sources / receiver. To further extract features, we utilize a vision transformer module~\cite{dosovitskiy2020image} \( F_{\text{vt}} \) that partitions each reflection-based map into patches, aggregates local features, and builds spatial dependencies among patches. This results in compact patch-level geometry representations: \( g'_{\text{r,rf}} = F_{\text{vt}}(I_{r, \text{rf}}) \), \( g'_{\text{s,rf}} = F_{\text{vt}}(I_{s,\text{rf}}) \) and \( \{ g'^{(k)}_{\text{ref,rf}} = F_{\text{vt}}(I^{(k)}_{\text{ref,rf}}) \}_{k=1}^K \), where each feature map has dimensionality \( N_p \times C_p \). Finally, we apply a MLP layer to project the patch dimension $N_p$ to 1, resulting in $g_{\text{r,rf}}$, $g_{\text{s,rf}}$, and $\{g^{(k)}_{\text{ref,rf}}\}^{K}_{k=1}$, respectively.

\subsubsection{Reference RIR Encoder}\label{reference_RIR_encoder}

To capture acoustic features related to energy decay and reverberation patterns within the room, we leverage reference RIRs as proxies for the acoustic characteristics at various source locations relative to the receiver. To encode these acoustic features, we first compute the log-magnitude spectrogram of each reference RIR using the Short-Time Fourier Transform (STFT): $\mathbf{S}_{\text{ref,k}} = \log(\|\text{STFT}(A_{\text{ref,k}})\|)$, where $\mathbf{S}_{\text{ref,k}} \in \mathbb{R}^{F \times T}$. To extract robust acoustic features, we implement the Reference RIR Encoder using ResNet-18, and use the mean pooled features \( f^{(k)}_{a} \in \mathbb{R}^d \) from the last layer to encode each reference RIR.

\subsubsection{Fusion and Weighting Module}\label{fusion_and_weighting}

The Fusion and Weighting Module integrates the outputs from the Geometric Feature Extractor and the Reference RIR Encoder to generate the target RIR prediction. This module combines geometric and acoustic features for reference sources as well as the geometric features of target source, finally computing the weights that are applied to reference RIRs.

\textbf{Fusion of Geometric and Acoustics Features.} For each reference source, we combine the geometric feature $g^{(k)}_{\text{ref,dir}}$,
$g^{(k)}_{\text{ref,rf}}$, $g_{\text{r,rf}}$ and the acoustic feature \( f_{a}^{(k)} \) by concatenating them along the feature dimension, resulting in: $\mathbf{h_{\textbf{ref}}^{(k)}} = \text{Concat}( g^{(k)}_{\text{ref,dir}}, g^{(k)}_{\text{ref,rf}}, g_{\text{r,rf}}, f_{a}^{(k)})$.

Similarly, for the target source, we combine the geometric feature $g_{\text{dir}}$, $g_{\text{s,rf}}$ and $g_{\text{r,rf}}$ via concatenation, yielding: $\mathbf{h'_{t}} = \text{Concat}(g_{\text{dir}}, g_{\text{s,rf}}, g_{\text{r,rf}})$. We then project 
the fused feature $\mathbf{h'_{t}}$ to $\mathbf{h_{t}}$ through a MLP to make the feature dimension the same as $\mathbf{h}_{\textbf{ref}}^{(k)}$.

To align the target and reference features, we compute the attention between the target fused vector \( \mathbf{h}_{t} \) and each reference fused vector \( \mathbf{h}^{(k)}_{\textbf{ref}} \). Specifically, given the reference fused features \( \mathbf{H_{\textbf{ref}}} = \{ \mathbf{h}^{(k)}_{\textbf{ref}} \}_{k=1}^K \) and the target fused vector \( \mathbf{h}_{t} \), the attention output \( \mathbf{Z} \) is computed as:

\[
\mathbf{Z} = \text{softmax}\left( \frac{\mathbf{H_{\textbf{ref}}} \cdot \mathbf{h^T_{t}}}{\sqrt{C}} \right) \odot  \mathbf{H_{\textbf{ref}}},
\]
where $\cdot$ and $\odot$ denote matrix multiplication and element-wise multiplication with broadcasting respectively, and $\mathbf{H_{\textbf{ref}}} \in \mathbb{R}^{K \times C}$, $\mathbf{h_{t}} \in \mathbb{R}^{1 \times C}$, $\mathbf{Z} \in \mathbb{R}^{K \times C} $. These attention outputs \(  \mathbf{Z} = \{ \mathbf{z}_{k} \}_{k=1}^K \) for each reference RIR is now attended by the fused feature of the target RIR.

\textbf{Time-Aligned Weighting Matrix.}
Given the attention outputs \( \mathbf{Z} \in \mathbb{R}^{K \times C} \), we next generate a time basis vector \( \mathbf{T}_{b} \) based on the temporal indices of the spectrogram \([0, 1, 2, \dots, T]\). Specifically, we compute \( \mathbf{T^{'}_{b}} \) using sinusoidal positional encoding~\cite{NIPS2017_3f5ee243} and then apply a MLP layer to project \( \mathbf{T^{'}_{b}} \) to \( C \), resulting in \( \mathbf{T_{b}} \in \mathbb{R}^{T \times C} \). We generate the time-aligned weighting matrix \( \mathbf{W} \in \mathbb{R}^{K \times T} \) by computing the outer product between \( \mathbf{Z} \) and \( \mathbf{T_{b}} \): \( \mathbf{W} = \mathbf{Z} \cdot \mathbf{T^T_{b}} \). Each row of \( \mathbf{W} \) corresponds to the weights applied to the log-magnitude spectrogram of each reference RIR, adapting them to match the temporal structure of the target spectrogram. This weighting matrix \( \mathbf{W} \) effectively shapes each reference spectrogram to align with the characteristics of the target RIR.

Finally, we predict the target RIR’s log-magnitude spectrogram \( \mathbf{S}_{\text{pred}} \) via the weighted sum of the log-magnitude spectrograms of the reference RIRs: \( \mathbf{S}_{\text{pred}} = \sum_{k=1}^K \mathbf{W}_{k} \odot \mathbf{S}_{\text{ref}, k} \). \( \mathbf{W}_{k} \) is the \( k \)-th row of the weight matrix \( \mathbf{W} \), applied to the corresponding log-magnitude spectrogram \( \mathbf{S}_{\text{ref}, k} \).

\subsubsection{Training and Inference} 
During training, we use the magnitude STFT \( L_1 \) Loss to compute the error between the magnitude spectrograms of the predicted target RIR and the ground-truth RIR: $\mathcal{L}_{\text{STFT}} = \|\exp(\mathbf{S}_{\text{pred}}) - \exp(\mathbf{S}_{\text{gt}}) \|_1$. Additionally, following \cite{majumder2022few}, we incorporate an energy decay loss to optimize the decay patterns of the predicted spectrogram. The energy decay loss \( \mathcal{L}_{\text{ED}} \) is defined as:
$\mathcal{L}_{\text{ED}} = \left\| \text{EDC}(\mathbf{S}_{\text{pred}}) - \text{EDC}(\mathbf{S}_{\text{gt}}) \right\|_1$,
where \( \text{EDC}(\cdot) \) denotes the energy decay curve of RIR in the frequency domain. The total loss becomes $\mathcal{L}_{\text{total}} = \mathcal{L}_{\text{STFT}} + \lambda \mathcal{L}_{\text{ED}}$, where \( \lambda \) is a weight to balance the contribution of the energy decay loss.

During inference, we randomly samples $K$ RIRs $\{A_{ref,k}\}^{k=K}_{k=1}$ along with corresponding source locations $\{P_{\text{ref}, k}\}^{k=K}_{k=1}$ from a test room as reference inputs. The model predicts the magnitude spectrogram of a target RIR, which is then converted back to a waveform via the Griffin-Lim~\cite{1164317} algorithm.

\section{The \ourdataset Dataset}
\label{sec:dataset}

To our best knowledge, there are two prior datasets with a large number of rooms: SoundSpaces MP3D~\cite{majumder2022few,chen2020soundspaces} and GWA~\cite{tang2022gwa}.
SoundSpaces MP3D comprises only 83 rooms with limited material variety (around 100 types), a fixed one-to-one mapping between semantic labels and acoustic coefficients, and a constrained 2D configuration at fixed heights. This setup restricts real-world applicability, as actual rooms often contain diverse materials with varying acoustic properties and require 3D spatial modeling to capture realistic sound propagation. 
For GWA, while it includes a large number of simulated RIRs from a wide variety of synthetic rooms and explicitly models wave propagation, the wave-based method it employs, PFFDTD~\cite{Hamilton2013ROOMAM}, is a lower-resolution approach. This method prioritizes computational efficiency, which comes at the cost of reduced simulation accuracy.

To address the above limitations, we introduce \ourdataset, a new large-scale, high-quality dataset of simulated RIRs specifically designed for robust generalization across diverse room geometries, sizes, and material properties. We use Treble Technology's simulation platform\footnote{https://www.treble.tech/}, where a more advanced wave-based solver, i.e., the Discontinuous Galerkin (DG) Method~\cite{osti_4491151}, is supported. Employing such techniques to simulate RIRs in our dataset is crucial for achieving cross-room generalization and sim-to-real transfer applications. \ourdataset simulates 260 rooms across 10 categories, featuring 300K simulated RIRs from different source-receiver pairs and full 3D spatial configurations. Each room includes a randomized material assignment from a library of 332 materials across 11 categories, ensuring diversity in acoustic properties even among similar geometries. The combination of scale, material diversity, and simulation fidelity enables \ourdataset to accurately reflect the acoustics of real-world environments.

\section{Experiments}
\label{sec:expt}

\begin{table*}[h!]
\small
\centering
\vspace{-2mm}
\begin{tabular}{lcccccc}
\toprule
\textbf{Method} & \multicolumn{3}{c}{\textbf{Seen Splits}} & \multicolumn{3}{c}{\textbf{Unseen Splits}} \\
\cmidrule(lr){2-4} \cmidrule(lr){5-7}
 & \textbf{EDT error (s)} & \textbf{C50 error (dB)} & \textbf{T60 error (\%)} & \textbf{EDT error (s)} & \textbf{C50 error (dB)} & \textbf{T60 error (\%)} \\
\midrule
Random Across Rooms & 0.290 & 6.831 & 37.35 & 0.313 & 7.802 & 35.15 \\
Random Same Room & 0.129 & 3.567 & 12.80 & 0.172 & 5.440 & 16.08 \\
Few-shot RIR~\cite{majumder2022few} (K=1) & 0.157 & 3.957 & 31.42 & 0.130 & 3.225 & 20.10 \\
Few-shot RIR~\cite{majumder2022few} (K=4) & 0.157 & 4.026 & 31.63 & 0.136 & 3.568 & 19.30 \\
Few-shot RIR~\cite{majumder2022few} (K=8) & 0.174 & 4.451 & 32.71 & 0.187 & 4.470 & 21.15 \\
Linear Interpolation (K=8) & 0.094 & 2.421 & 9.76 & 0.121 & 3.090 & 13.73 \\
Nearest Neighbor (K=8) & 0.064 & 1.717 & 8.94 & 0.090 & 2.667 & 11.64 \\
\ourapproach{} (K=1) & 0.046 & 1.183 & 9.50 & 0.075 & 1.841 & 13.47 \\
\ourapproach{} (K=4) & 0.040 & 1.005 & 8.15 & 0.068 & \textbf{1.335} & 13.28 \\
\ourapproach{} (K=8) & \textbf{0.038} & \textbf{0.940} & \textbf{8.13} & \textbf{0.055} & 1.457 & \textbf{10.53} \\
\bottomrule
\end{tabular}
\vspace{-2mm}
\caption{\textbf{Cross-Room RIR Prediction Results for Both the Seen and Unseen Splits.}  We report EDT Error (EDT) in seconds, C50 Error (C50) in dB, and T60 percentage error (T60), with lower values indicating better performance. For Few-shot RIR~\cite{majumder2022few} and \ourapproach (Ours), we evaluate in a few-shot manner by setting the number of reference RIRs $K$ to 1, 4, and 8.}
\label{tab:eval_combined_cross}
\end{table*}

\begin{table*}[h!]
\centering
\small
\vspace{-1.5mm}
\begin{tabular}{lcccccccccccc}
\toprule
\textbf{Method} & \multicolumn{3}{c}{\textbf{Classroom}} & \multicolumn{3}{c}{\textbf{Dampened Room}} & \multicolumn{3}{c}{\textbf{Hallway}} & \multicolumn{3}{c}{\textbf{Complex Room}} \\
\cmidrule(r){2-4} \cmidrule(r){5-7} \cmidrule(r){8-10} \cmidrule(r){11-13}
& \textbf{EDT} & \textbf{C50} & \textbf{T60} & \textbf{EDT} & \textbf{C50} & \textbf{T60} & \textbf{EDT} & \textbf{C50} & \textbf{T60} & \textbf{EDT} & \textbf{C50} & \textbf{T60} \\
\midrule
Random Across Rooms & 0.546 & 8.740 & 19.03 & 0.771 & 18.726 & - & 0.874 & 11.025 & 21.71 & 0.472 & 7.392 & 16.01 \\
Random Same Room & 0.160 & 3.092 & 3.12 & 0.099 & 6.840 & - & 0.308 & 6.461 & 16.61 & 0.218 & 4.566 & 5.66 \\
Linear Interpolation (K=8) & 0.113 & 2.172 & 4.42 & 0.058 & 4.584 & - & 0.088 & 2.127 & 4.55 & 0.124 & 2.848 & 5.17 \\
Nearest Neighbor (K=8)  & 0.108 & 1.949 & \textbf{2.71} & \textbf{0.044} & \textbf{3.278} & - & 0.068 & 0.990 & \textbf{3.02} & 0.091 & 1.936 & \textbf{2.53} \\
Diff-RIR~\cite{wang2024hearing} (K=12) & 0.113 & 2.147 & 12.39 & 0.100 & 3.796 & - & 0.160 & 2.049 & 14.34 & 0.115 & 2.027 & 12.76 \\
\ourapproach (K=8) (Ours) & \textbf{0.093} & \textbf{1.628} & 6.25 & 0.044 & 3.302 & - & \textbf{0.062} & \textbf{0.954} & 3.20 & \textbf{0.077} & \textbf{1.688} & 4.33 \\
\bottomrule
\end{tabular}
\caption{\textbf{Sim-to-Real Transfer Results in Four Real Environments from the Hearing-Anything-Anywhere Dataset~\cite{wang2024hearing}.} We report EDT Error (EDT) in seconds, C50 Error (C50) in dB, and T60 percentage error (T60). Due to noisy measurements in the dampened room, resulting in low SNR and invalid T60 calculations on the EDC curve, we omit this metric for the dampened room.}
\label{tab:sim_to_real}
\end{table*}

\subsection{Implementation Details} 

In the \ourdataset dataset, RIRs are sampled at 22,050 Hz with a maximum length of 9600 samples (0.435 s). We compute the magnitude spectrogram \(S\) with FFT size 124, window size 62, and hop size 31, yielding a shape of \(63 \times 310\). Panorama depth maps of room geometry are rendered at a resolution of \(256 \times 512\) from the receiver’s location, and source/receiver positions are recorded as 3D coordinates \((x, y, z)\). For \ourapproach, we implement a Vision Transformer block \(F_{\text{vt}}\) with 6 multi-head attention layers (8 heads, hidden size 512). Depth maps are divided into \(16 \times 32\) patches, resulting in all reflection-based features such as \(g_{\text{r,rf}}\) and \(g_{\text{s,rf}}\) of dimension \(256 \times 512\). Direct path features are calculated using sinusoidal encoding on each 3D coordinate with 20 frequency bins, and are then projected into 256-dimensional vectors via MLP. For loss calculation, we set \(\lambda = 0.01\) to balance the STFT loss and the energy decay loss.

\subsection{Baselines}

We compare with a series of baselines as well as prior methods~\cite{majumder2022few,wang2024hearing}:

\begin{itemize}
    \item \textbf{Random Across Rooms}: Randomly sample a RIR from the entire dataset as the prediction for the target RIR.

    \item \textbf{Random Same Room}: Randomly sample a RIR from the same room as as the prediction for the target RIR.

    \item \textbf{Nearest Neighbor}: Sample \( k \)-shot reference RIRs and select the RIR with the closest spatial distance to the target source as the prediction.

    \item \textbf{Linear Interpolation}: Linearly interpolate between \( k \)-shot reference RIRs based on the distance between each reference and the target source location.

    \item \textbf{Few-Shot RIR~\cite{majumder2022few}}: Few-Shot RIR implements a transformer architecture that fuses features from separate encoders of multi-modal conditional inputs and then generates the target RIR by decoding the transformer outputs via a UNet decoder. We adapt their model to our task by replacing the binaural echos with our single-channel reference RIRs (different source and receiver locations) and using panorama depth images as inputs to the image encoder instead of egocentric RGBD images. 

    \item \textbf{Diff-RIR~\cite{wang2024hearing}}: We compare with Diff-RIR in evaluation on sim-to-real transfer. The framework utilizes the few-shot, i.e., 12 reference RIRs, to train a differentiable rendering pipeline to learn acoustics parameters of the room geometry. For fair comparison, we finetune our \ourapproach model pre-trained on \ourdataset on the same set of reference RIRs as Diff-RIR in each room, and then test on the same test split. Note that Diff-RIR requires training one model per each room and the training process becomes computationally infeasible for large space with complex room geometries. Therefore, we do not include it in our comparison on \ourdataset.
\end{itemize}

 In addition, for a more complete comparison with prior methods on RIR prediction, we also adapt our method for the single-room RIR prediction task and compare with prior work~\cite{su2022inras, luo2022learning}. Please see Supp. for results.

\subsection{Metrics} We evaluate the energy pattern of the generated RIRs against ground-truth RIRs using three key acoustic metrics, which are strongly correlated with hearing perception and commonly used in prior work on RIR prediction~\cite{su2022inras,luo2022learning}:
\begin{itemize}    
    \item \textbf{Early Decay Time (EDT) Error:} To evaluate early reflection characteristics, we use the EDT error, which measures the time taken for the initial \( 5 \, \text{dB} \) decay in the energy curve.
    
    \item \textbf{Clarity (\( \mathbf{C50} \)):} For comparing early-to-late energy ratios, we employ the clarity metric \( C50 \), which provides insights into the prominence of early reflections over later reverberations.

    \item \textbf{\( \mathbf{T60} \) Error:} We evaluate the accuracy of reverberation time by comparing the \( T60 \) value of the predicted RIR and the ground-truth RIR. We calculate \( T60 \) using \( T20 \), based on a linear fit between \(-5 \, \text{dB}\) and \(-25 \, \text{dB}\) on the logarithmic energy decay curve obtained from Schroeder Backward Integration~\cite{schroeder1965new}.
\end{itemize}

\begin{figure*}[tp]
    \centering
    \includegraphics[width=\linewidth]{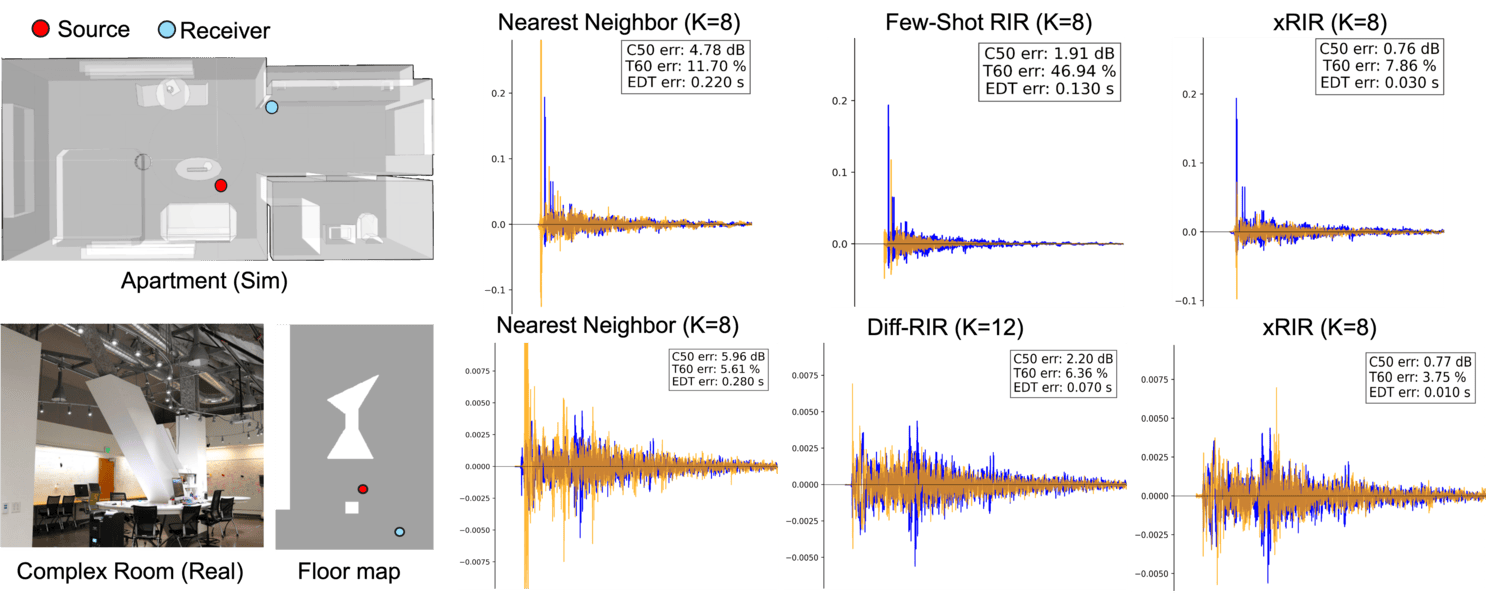}
        \vspace{-0.05in}
    \caption{\textbf{Qualitative Comparisons of RIR Predictions.} We compare the performance of our method and the baselines both in simulated (top row) and real (bottom row) environments. Room geometry, sample RIR predictions, and the corresponding error metrics are included. \ourapproach shows more accurate RIR predictions in both settings.
    }
    \label{fig:qual_results}
\end{figure*}

\begin{figure*}[tp]
    \centering
    \includegraphics[width=0.99\linewidth]{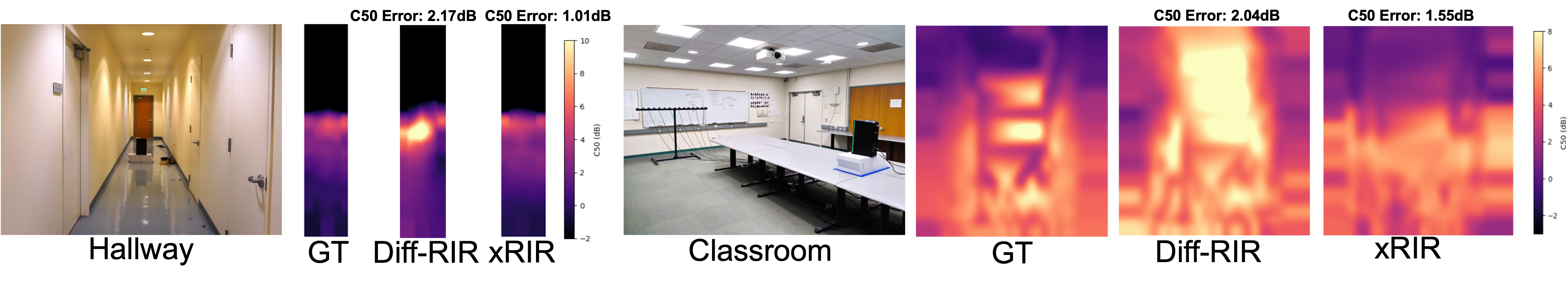}
    \vspace{-0.05in}
    \caption{\textbf{Qualitative Comparisons of Acoustic Map Predictions in Two Real Environments: a Hallway and a Classroom.} We visualize the acoustics maps by computing the C50 metric at dense locations in the entire room and compare with the ground-truth acoustic map. \ourapproach achieves C50 distributions that better matches the ground-truth.}
    \label{fig:qual_results_2}
\end{figure*}

\subsection{Quantatitive Results on \textbf{\ourdataset}}

We show cross-room RIR prediction results in both environments seen during training as well as unseen new environments. As shown in Table \ref{tab:eval_combined_cross}, our \ourapproach model significantly outperforms all baselines across all metrics (EDT error, \(C50\) error, and \(T60\) error). In the seen split, our \ourapproach model yields the lowest errors across metrics, particularly in \(C50\) and \(T60\). Our gains persists in the unseen split. In particular, our model with \(K=8\) reduces \(T60\) error to around 10\%, while other baselines exhibit much higher errors. This result highlights our model's robustness in capturing reverberation characteristics across different room configurations and the ability to generalize to unseen environments with different room acoustic properties.

The Few-Shot RIR approach from~\cite{majumder2022few} does not perform well \ourdataset. We suspect that this is due to two factors: i) their UNet decoder struggles to reconstruct high-fidelity RIRs on our data, as it relies on highly compressed fusion features; ii) their method uses binaural echoes with co-located source and receiver positions, which fundamentally differ from our setup, where reference RIRs are measured with the source and receiver at different locations. This spatial disparity likely impacts feature relevance, limiting its performance on our dataset.
 
\subsection{Sim-to-Real Transfer to Real Environments} 

To evaluate whether our model can also generalize to real-world environments, we use four real rooms from the Hearing-Anything-Anywhere Dataset~\cite{wang2024hearing}. We compare our method against Diff-RIR~\cite{wang2024hearing}, a physics-based differentiable RIR rendering pipeline that utilizes 12 reference RIRs per room to predict RIRs for new locations. As shown in Table \ref{tab:sim_to_real}, our model compares favorably against all baselines. In partilar, despite using only 8-shot references, our method outperforms Diff-RIR that uses 12 reference RIRs in all acoustic metrics, demonstrating its strong generalization capabilities. We observe that our method underperforms on the T60 metric compared to the Nearest Neighbor baseline across all four rooms. We suspect this is because T60, as a global metric, is more sensitive to measurement noise due to its aggregation of all acoustic interactions within the room. Our learning-based method can struggle with low SNR beyond the early parts of the waveform, as it is trained on simulation data with higher SNR than real room measurements. In contrast, EDT focuses on early reflections with high SNR, make it less noise-sensity, and C50 is similarly robust due to noise smoothing in the integration beyond the early parts. Despite this, our results demonstrate that \ourapproach's effectiveness in adapting from simulated rooms to real environments, successfully capturing diverse room acoustics with fewer reference RIRs than prior methods.

\subsection{Qualitative Results} 

 We present qualitative results by comparing the predicted RIRs and acoustic maps between our model \ourapproach and the baseline methods, in both simulated and real environments.
 
\textbf{RIR Predictions.} In Fig.~\ref{fig:qual_results}, we visualize sample results of RIR waveforms on a simulated apartment and a real room with complex geometry. Side-by-side comparison shows that predicted RIRs from \ourapproach align more closely with the ground-truth RIR waveforms in the early part than baselines. This observation is consistent with the low acoustics metrics errors achieved by our method in the quantitative results shown in Table \ref{tab:eval_combined_cross}.
 
\textbf{Acoustics Maps.} Furthermore, we compute the RIRs at dense locations across the entire real rooms, and compute the clarity of the predicted and ground-truth RIRs to reconstruct acoustics maps according to the floor plans. As shown in Fig.~\ref{fig:qual_results_2}, across dense locations in these rooms, overall \ourapproach achieves better C50 distribution than Diff-RIR~\cite{wang2024hearing} compared to the ground-truth acoustic maps, especially at moderate-to-low intensity regions. These qualitative results demonstrate the effectiveness of \ourapproach in accurate RIR prediction in both simulation and real-world settings.

\vspace{-0.1in}
\section{Conclusion}
\label{sec:conclude}

We presented \ourapproach, a model designed for generalizable RIR prediction across diverse room environments. To tackle the cross-room RIR prediction task, we also introduced a large-scale, realistic RIR simulation dataset, \textsc{AcousticRooms}\xspace, which includes diverse room categories, geometries, and material properties. Results under the simulation settings show that our framework outperforms prior methods and strong baselines in both seen and unseen environments. Furthermore, sim-to-real transfer experiments reveal that our model, pre-trained on simulated data, effectively generalizes to real-world settings. Future work may focus on improving modeling techniques, such as using generative approach as proposed for sound generation~\cite{NEURIPS2024_b782a346,su2024vision} to achieve better performance on acoustic modeling with minimal reference RIRs, or dynamically choose the suitable number of reference RIRs needed depending on the complexity of the environment.

{
    \small
    \bibliographystyle{ieeenat_fullname}
    \bibliography{main}
}

\clearpage
\setcounter{page}{1}
\maketitlesupplementary

\section{Summary of Supplementary Materials}
\noindent In our supplementary materials, we provide:
\begin{enumerate}
    \item Supplementary video. \ref{sec:supp_video}
    \item Additional details including explanation of reference RIR setup, implementations of \ourapproach and baselines as well as experiments setup, see Section \ref{sec:additional_implementation_details}.
    \item More details about \ourdataset, see Section \ref{sec:dataset_detail}.
    \item Single-room RIR prediction results compared to prior works, see Section \ref{sec:single_room_expt}.
    \item  Additional experiments and ablation studies of the xRIR, see Section \ref{sec:ablation}.
    \item Additional qualitative samples, RIR prediction comparison \ref{sec:qual_results_suppl}.
\end{enumerate}

\section{Supplementary Video}
\label{sec:supp_video}
In the supplementary video, we provide a brief summary of our work and qualitative samples of audio rendered with different predicted RIRs in simulation as well as real environments. For best perceptual experience, please turn the Audio ON and use headphone when watching. 

For demos in the supplementary video that demonstrate audio rendering along trajectories in real scenes, we convolve our predicted single-channel Room Impulse Response (RIR) at each point in the trajectory with a Head-Related Impulse Response (HRIR) from a predefined Head-Related Transfer Function (HRTF). This process yields binaural RIRs, which capture spatial effects. We then convolve these binaural RIRs with the source audio to obtain the binaural audio along these trajectories.

\section{Additional Details}
\label{sec:additional_implementation_details}

\textbf{Further explanation on reference RIR setup:} Our reference RIRs setup is practically useful. Model trained on this setup is able to predict new RIRs without any \textit{reference RIRs re-measurement} when either source or receiver freely moves in the target room. Our method is capable of addressing both scenarios below:

i) \textit{Fixed reference receiver, multiple sources}: Measure reference RIRs between a fixed receiver and sources at various locations, then the model predicts the RIR for the fixed receiver and a new source location (\textsc{AcousticRooms} setup).

ii) \textit{Fixed reference source, multiple receivers}: Measure reference RIRs between a fixed source and receivers at various locations, then the model predicts the RIR for the fixed source and a new receiver location (HearAnythingAnywhere setup).

A model trained on i) can be directly applied to ii) by switching source and receiver subscripts (due to symmetry of wave equation for single-channel RIR), as shown in our sim-to-real transfer experiment. We adopt scenario i) in our task, where the receiver always matches the reference receiver location.

\textbf{\ourapproach:} For \ourapproach, we implement a Vision Transformer block \(F_{\text{vt}}\) with 6 multi-head attention layers (8 heads, hidden size 512). For panorama depth map, the center pixel corresponds to the receiver location. Two spherical angles maps are initialized for equirectangular projection from the depth map to 3D coordinates map. In the vision transformer module, the 3D coordinates map is divided into \(16 \times 32\) patches, resulting in all reflection-based features such as \(g_{\text{r,rf}}\) and \(g_{\text{s,rf}}\) of dimension \(256 \times 512\). Direct path features are calculated using sinusoidal positional encoding on each 3D coordinate with 20 frequency bins, and are then projected into 256-dimensional vectors via MLP. Similarly, the time basis vector $T_b$ is calculated by sinusoidal positional encoding with 10 frequency bins for each time index, where the length of $T_b$ is same as the length of spectrogram, 310. Before performing weight combination, we further preprocess the reference RIRs by time-shifting them based on the distance difference between the target and reference source-receiver pairs, divided by the speed of sound. For loss calculation, we set \(\lambda = 0.01\) to balance the STFT loss and the energy decay loss.

\vspace{0.05in}

\noindent \textbf{Few-Shot RIR:} Unlike the approach and the problem setup in ~\cite{majumder2022few}, which use binaural echoes where the source and receiver are co-located to predict a target binaural RIR, we use reference RIRs measured with the source at different locations from the receiver location to predict single-channel RIR at a target source. This is very important since the echo input used by ~\cite{majumder2022few} are infeasible to obtain under the single-channel RIR scenario, because it is not reasonable in physics to co-locate source and receiver at the same location to measure the single-channel RIR. In addition, we also omit the RGB image for the visual input and use only depth maps as inputs to the vision branch of the Few-shot RIR model, due to the weak correspondence between room semantics and material properties in \ourdataset dataset. We also emphasize that we use panorama depth images captured from each reference source location instead of egocentric depth images as the depth inputs. For all depth observations, we rendered at a resolution of $128 \times 256$. Except for the adaptations above, all other implementation details follow the Few-Shot RIR model~\cite{majumder2022few}.

\vspace{0.05in}

\noindent \textbf{Diff-RIR:} We use their released Github code and model checkpoints to perform evaluations on all rooms in the Hearing-Anything-Anywhere Dataset~\cite{wang2024hearing}. We strictly followed the inference and evaluation settings in the paper, and obtained the same results in terms of the metric errors reported in the paper (Mag and Env error metrics) to make sure there are no implementation issues. And then we evaluate their inference results on our three acoustic metrics which are more related to perceptual quality of the RIRs: EDT error, C50 error and T60 error.

\vspace{0.05in}

\noindent  \textbf{Experiments Setup:} For cross-room RIR prediction experiments on \ourdataset, we manually select 10 sources in each simulated room as candidate reference sources to make sure that their spatial locations are evenly distributed within the scene as much as possible. For the seen setting, we split training and test set within each room by receivers, where the RIRs of 90\% of receivers belong to training split and remaining 10\% belong to test split. For the unseen setting, we split the data by rooms. For each room category, we use 90\% of rooms for training and 10\% rooms for testing. During both training and testing, we randomly select $K = 1, 4, 8$ reference RIRs from these candidate sources. And for each $K$, we train a separate model on the dataset.\\
For experiments on the Hearing-Anything-Anywhere Dataset~\cite{wang2024hearing}, we sample reference $K = 8$ RIRs from their selected 12 reference RIRs in each room. In this real-world dataset, the source has specific directivity patterns which are not captured in our pretrained model using \ourdataset. Therefore, we further finetune the pretrained \ourapproach model on these 12 selected RIRs to make sure a fair comparison with Diff-RIR ~\cite{wang2024hearing}. Specifically, in each iteration during finetuning, we randomly sample 8 of 12 as reference RIRs and predict a target RIR sampled from remaining 4 RIRs. We also use the same validation set as Diff-RIR to select the model checkpoint for testing. We select the one with lowest validation loss to evaluate on the test split of the dataset. It is note-worthy that even though \ourapproach is finetuned, compared to the training time of Diff-RIR on these few shot samples (6 hours / scene), \ourapproach converges much faster than Diff-RIR, with a matter of minutes. This helps us to quickly perform sim-to-real transfer across different environments efficiently.

\begin{figure*}[tp]
    \centering
    \includegraphics[width=\linewidth]{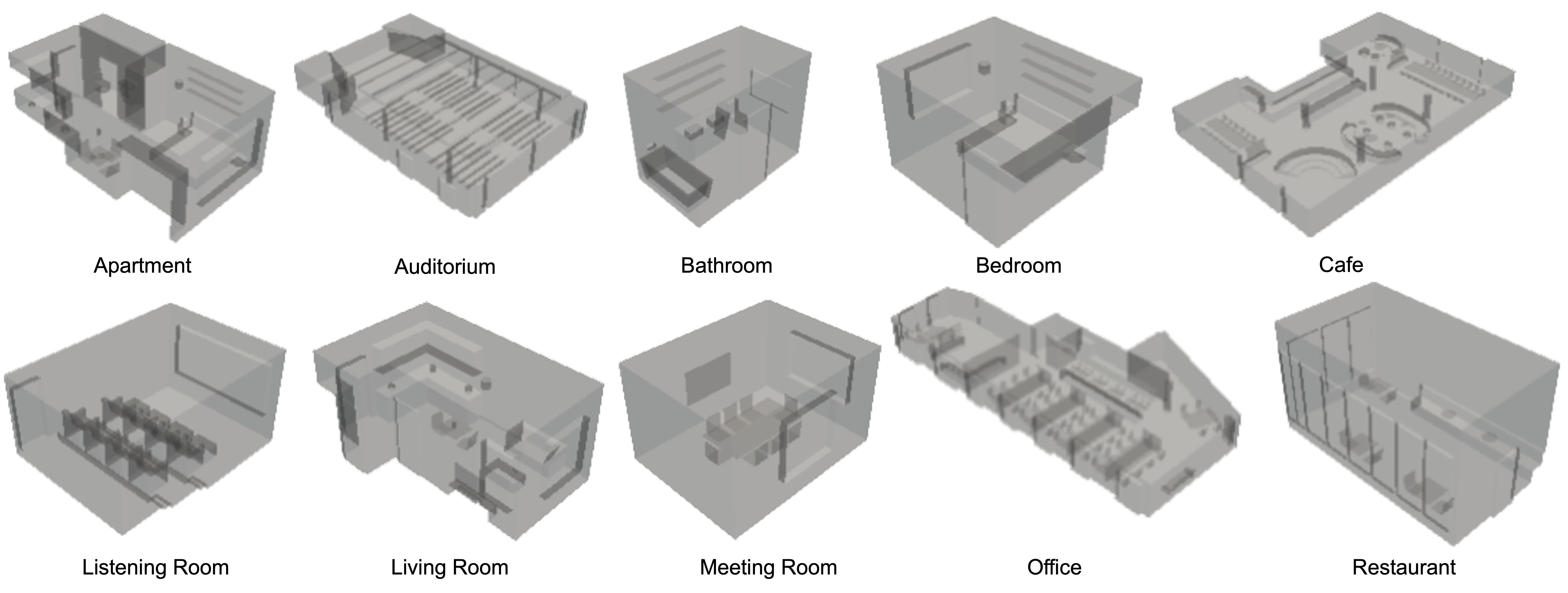}
    \caption{A visualization of different room categories in \ourdataset.}
    \label{fig:room_category}
\end{figure*}

\begin{figure*}[tp]
    \centering
    \includegraphics[width=\linewidth]{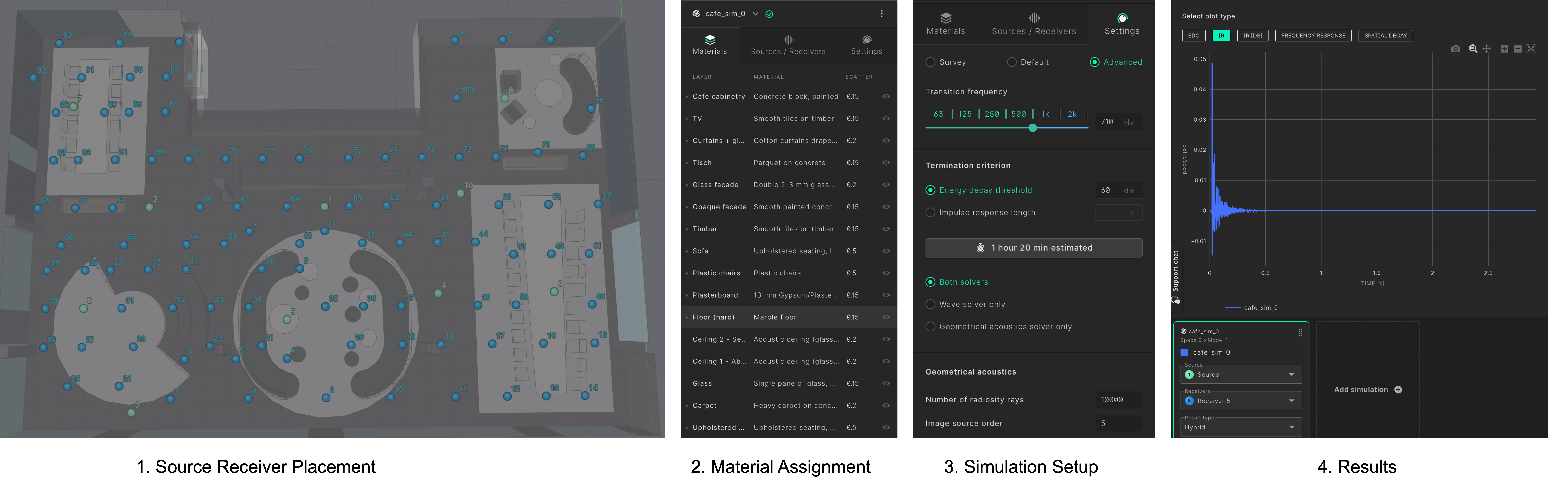}
    \caption{An overview of procedures to simulate RIRs in \ourdataset.}
    \label{fig:sim_setup}
\end{figure*}

\section{Dataset Details}
\label{sec:dataset_detail}
In this section, we present details about our large-scale simulated RIR dataset, \ourdataset, used for cross-room RIR prediction task. \ourdataset contains 260 rooms from 10 different categories, simulating a total of 30,000 RIRs from different source-receiver pairs. \ourdataset features professional room architecture designs of high quality, covering a wide range of room categories, including: apartment, auditorium, bathroom, bedroom, cafe, listening room, living room, meeting room, office and restaurant, as shown in Figure \ref{fig:room_category}. The area of rooms ranges from $20m^3$ to $1000m^3$, with diverse range of sizes and geometries. \ourdataset uses a realistic and commercial acoustics simulation platform, Treble, to perform single-channel RIR simulation. The platform supports a wide variety of simulation methods along with specific settings. To obtain more realistic RIR data as well as simulate large-scale data, we adopt the hybrid-based simulation. At low frequency bands, an advanced wave-based method is used to capture more subtle wave interaction effects such as diffraction and resonance. At high frequency bands, we use geometric-based simulation that combines two simulation techniques: image-source technique and stochastic ray-tracing technique.

\vspace{0.05in}

\noindent  \textbf{Source Receiver Placement:} To set up the simulation for each room, we first choose the sources and receivers and place them at different locations. For sources, we use omni-directional source devices without particular directivity patterns since our goal is not to overfit the model to a particular device pattern. Similarly, for receivers, we use monaural receivers such that they do not model specific HRTF patterns. Depending on the size of rooms, we place 10 to 100 sources and 25 to 100 receivers per room, to ensure they sufficiently cover the whole area of the rooms. To determine the location of each source and receiver, we apply a set of placement rules to avoid interference among devices and room surfaces when the distance becomes too small to cause issue in the simulation quality. We require that: i) Sources should be at least $0.5m$ away from each room surface, $1.0m$ away from other sources and $1.0m$ away from receivers. ii) Receivers should be at least $0.5m$ away from room surface and at least $0.5m$ away from the sources. Given these rules, we apply a point-picking algorithm to randomly sample valid source and receiver locations within each room at different height level from $0.5m$ to $2.5m$. 

\vspace{0.05in}

\noindent  \textbf{Material Assignment:} Once the source and receivers are determined, we assign materials to room surfaces by associating their semantic labels with particular material category. Treble platform provides a large-scale material database with 332 specific materials from 11 material categories, with each category containing 30 different material coefficients on average. We define the mapping between each semantic labels of room surfaces and the 11 material categories. In each room, each semantic label of a particular surface gets randomly mapped to one of the specific material with a set of acoustics coefficients under the material category. Different from existing RIR datasets~\cite{chen2020soundspaces, chen2022soundspaces, tang2022gwa}, this random assignment ensures enough diversity in the material properties of the room surfaces. Even two rooms share similar geometries and semantic objects, they could have very different acoustics behavior due to differences in their specific acoustics coefficients of materials.

\vspace{0.05in}

\noindent  \textbf{Simulation Setup:} Once sources and receivers are in place and room materials get assigned, we set up the simulation by specifying the hybrid mode to split the geometric-based and wave-based method.  We choose the crossover frequency to be $f_{\text{cross}} = 710 \text{Hz}$ between two methods such that wave-based effects could be sufficiently captured across different rooms and objects of different size. For geometric-based simulation, we use image source method up to reflection order of 4 with $50k$ rays emitting from the source. For refection orders higher than 4, we apply stochastic ray-tracing method with 5000 rays. With this configuration, we simulate the RIR of each room until $60dB$ energy decay to ensure all possible acoustics effects are sufficiently captured.

\begin{table*}[th!]
\small
\centering

\begin{tabular}{lcccccccccccc}
\toprule
\multirow{2}{*}{\textbf{Model}} & \multicolumn{3}{c}{\textbf{Apartment 1}} & \multicolumn{3}{c}{\textbf{Apartment 2}} & \multicolumn{3}{c}{\textbf{FRL Apartment 2}} \\
\cmidrule(lr){2-4} \cmidrule(lr){5-7} \cmidrule(lr){8-10}
 & EDT$\downarrow$ & C50 $\downarrow$ & T60 $\downarrow$ & EDT $\downarrow$ & C50 $\downarrow$ & T60 $\downarrow$ & EDT $\downarrow$ & C50 $\downarrow$ & T60 $\downarrow$ \\
\midrule
NAF & 0.077 & 0.426 & 7.508 & 0.066 & 0.453 & 7.925 & 0.088 & 0.420 & 6.308 \\
INRAS & 0.027 & 1.036 & 6.514 & 0.025 & 0.843 & 5.816 & 0.022 & 0.634 & 2.224 \\
xRIR & 0.026 & 1.000 & 6.192 & 0.031 & 0.932 & 5.755 & 0.021 & 0.587 & 1.972 \\
\bottomrule
\end{tabular}

\vspace{1em}

\begin{tabular}{lcccccccccccc}
\toprule
\multirow{2}{*}{\textbf{Model}} & \multicolumn{3}{c}{\textbf{Room 2}} & \multicolumn{3}{c}{\textbf{FRL Apartment 4}} & \multicolumn{3}{c}{\textbf{Office 4}} & \multicolumn{3}{c}{\textbf{Mean}} \\
\cmidrule(lr){2-4} \cmidrule(lr){5-7} \cmidrule(lr){8-10} \cmidrule(lr){11-13}
 & EDT $\downarrow$ & C50 $\downarrow$ & T60 $\downarrow$ & EDT $\downarrow$ & C50 $\downarrow$ & T60 $\downarrow$ & EDT $\downarrow$ & C50 $\downarrow$ & T60 $\downarrow$ & EDT $\downarrow$ & C50 $\downarrow$ & T60 $\downarrow$ \\
\midrule
NAF & 0.056 & 0.407 & 4.969 & 0.085 & 0.421 & 7.475 & 0.081 & 0.337 & 6.760 & 0.076 & \bf{0.411} & 6.824 \\
INRAS & 0.020 & 0.555 & 1.990 & 0.022 & 0.625 & 2.145 & 0.014 & 0.610 & 3.251 & 0.022 & 0.717 & 3.657 \\
xRIR & 0.019 & 0.541 & 1.910 & 0.021 & 0.561 & 2.140 & 0.013 & 0.502 & 2.767 & \bf{0.022} & 0.687 & \bf{3.456} \\
\bottomrule
\end{tabular}
\caption{Performance comparison on single-room RIR prediction task on six rooms in SoundSpaces 1.0 - Replica dataset in terms of EDT (s), C50 (dB), and T60 (\%) error metrics. }
\label{tab:ir_interp_result}
\end{table*}

\section{Comparison on Single-Room RIR Prediction}
\label{sec:single_room_expt}

\vspace{0.05in}

\noindent  \textbf{Adaptation of \ourapproach:} Although \ourapproach focuses on solving cross-room RIR prediction task, it could be easily adapted to single-room RIR prediction task as well. By removing all the components related to reference RIRs, we extract the geometric and spatial features related to only target source and receiver positions. We then follow ~\cite{su2022inras} to use their implicit neural decoder to perform RIR waveform synthesis. Specifically, given a time basis vector $\mathbf{B}$, and the outputs from Geometric feature extractor $g_{dir}$, $g_{r, \text{rf}}$ and $g_{s,\text{rf}}$,  we learn a implicit neural mapping function to synthesize time domain RIR waveform from the outer product between $\mathbf{B}$ with the three features: $\hat{A}_{t} = F_{\text{inr}} (g_{dir} \mathbf{B^T}, g_{r, \text{rf}} \mathbf{B^T}, g_{s,\text{rf}} \mathbf{B^T})$. We train the adapted model with same loss function as in ~\cite{su2022inras}, the multi-resolution STFT loss combined with waveform L2 loss.

\vspace{0.05in}

\noindent  \textbf{Experiment Setup:} In single-room RIR prediction task, the goal is to the fit scene acoustic with dense RIR observations. Therefore, we use the standard dense RIR dataset, SoundSpaces 1.0 Replica, to perform the experiment. Following prior works ~\cite{luo2022learning,su2022inras}, we use the six scenes from Replica. But instead of using binaural RIR data, we use single-channel RIR data by extracting the first channel of ambisonic RIR data of these scenes.
For each scene, we split the RIR data into training and test set with a ratio of 9:1. We cut the RIR to the maximum length of 8000 samples (0.363s) at sampling rate 22,050Hz for all six rooms. And for panorama image at each receiver location, we render it by setting the orientation to 0 in the habitat simulator~\cite{savva2019habitat}.

\vspace{0.05in}

\noindent  \textbf{Baselines:} We compare \ourapproach with two prior works of the state-of-the-art performance on the dataset, NAF~\cite{luo2022learning} and INRAS ~\cite{su2022inras}. For both methods, we remove the orientation conditioning vector to adjust for single-channel RIR prediction, while keeping the remaining implementations the same. For evaluation, we use the same metrics in the cross-room RIR prediction task for comparison. 

\vspace{0.05in}

\noindent  \textbf{Quantatitive Results:} We report individual results for each of six scenes and their average results and show them in Table \ref{tab:ir_interp_result}. As could be seen, when compared to NAF, \ourapproach outperforms on both EDT and T60 metrics, while slightly underperforming in C50 error. When compared to INRAS, \ourapproach outperform INRAS on 5 out of 6 scenes, except ``Apartment 2''. The reason is due to the fact that the mesh of ``Apartment 2'' shows significant amount of holes in a region, which leads to degraded quality of panorama depth inputs. While, INRAS does not suffer from this degradation since the method samples mesh points instead of rendering images. Overall, \ourapproach, when adapted to single-room RIR prediction task, shows on par or even better performance when compared to these prior arts, demonstrating the effectiveness of Geometric Feature Extractor to learn the scene acoustics from local geometric observations (around receiver).

\begin{figure*}[tp]
    \centering
    \includegraphics[width=\linewidth]{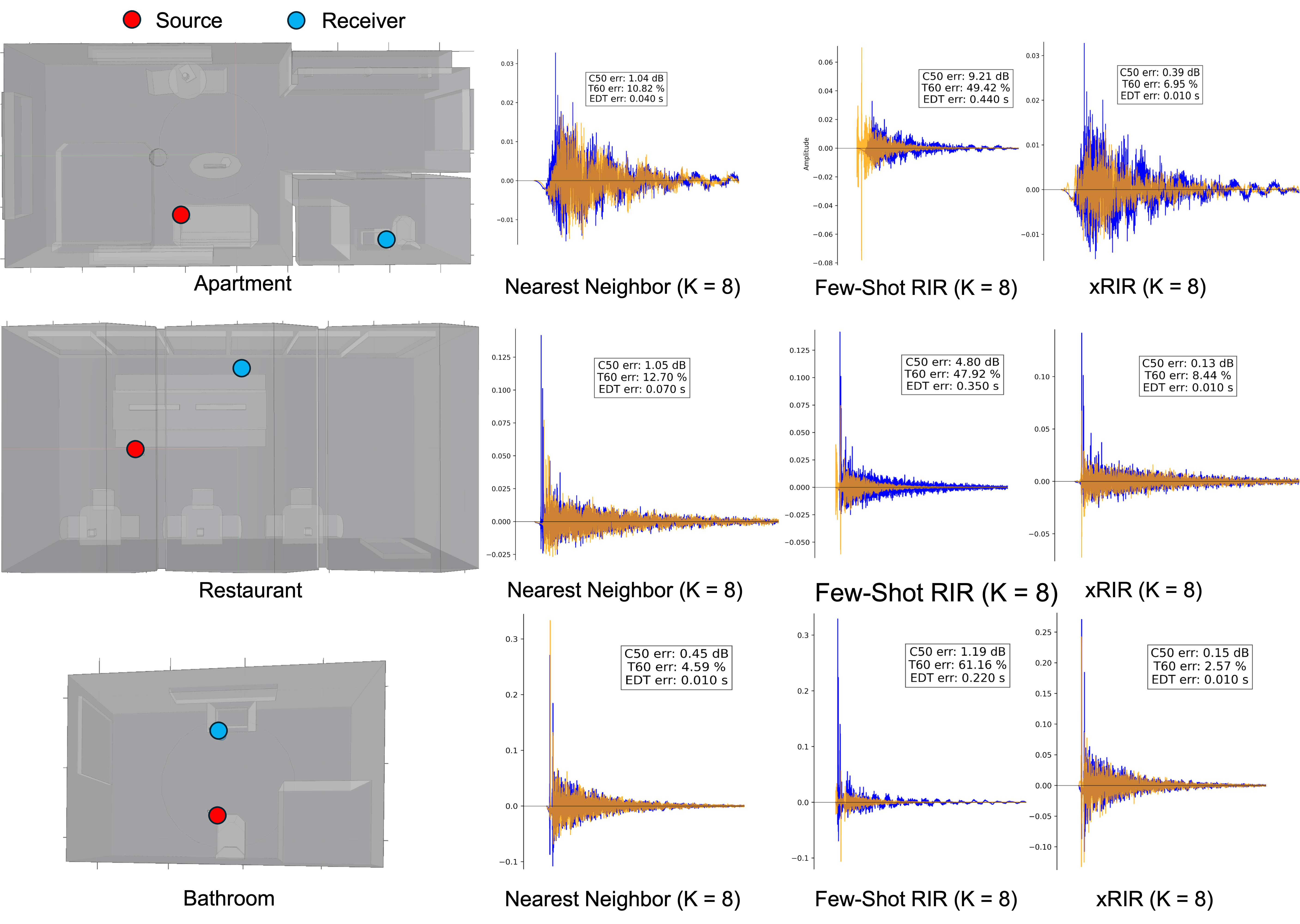}
    \vspace{-0.2in}
    \caption{Additional qualitative comparisons on RIR waveform predictions in \ourdataset.}
    \vspace{-3mm}
    \label{fig:add_qual_rir_1}
\end{figure*}

\begin{figure*}[tp]
    \centering
    \includegraphics[width=\linewidth]{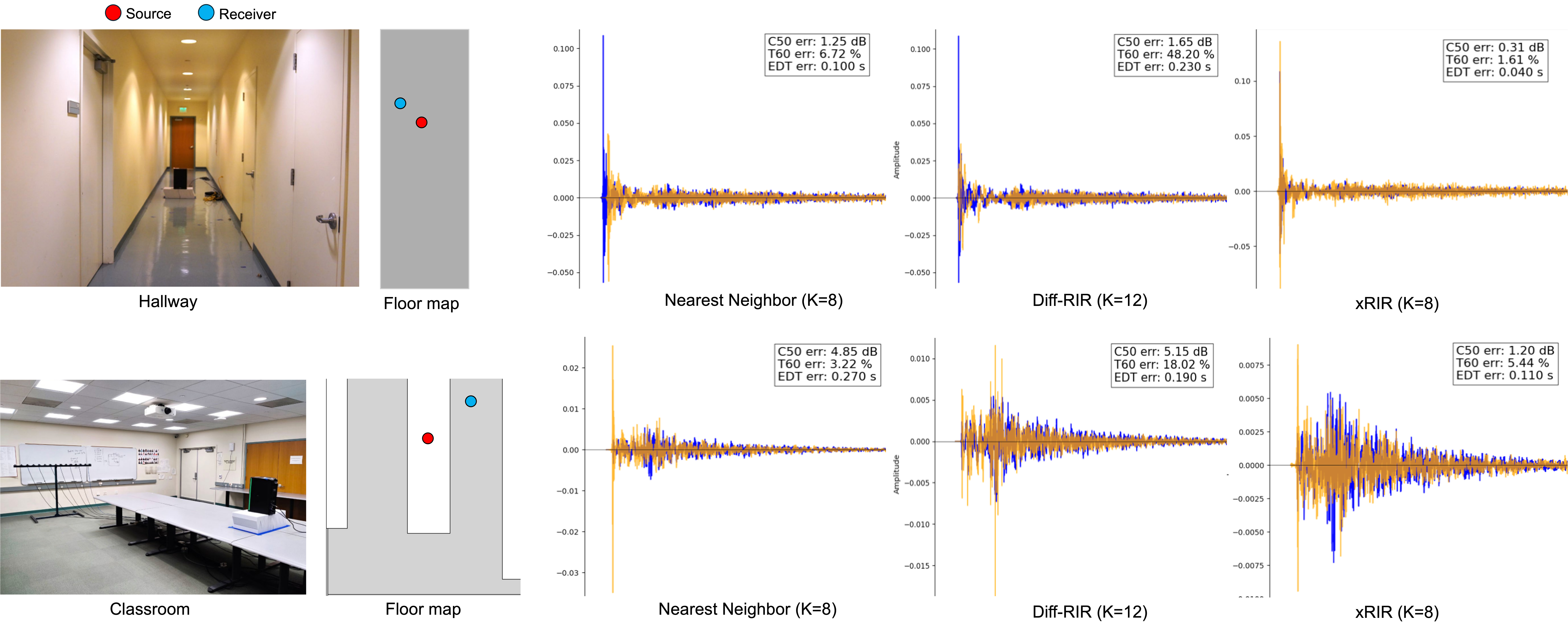}
    \vspace{-0.2in}
    \caption{Additional qualitative comparisons on RIR waveform predictions on the Hearing-Anything-Anywhere Dataset.}
    \vspace{-3mm}
    \label{fig:add_qual_rir_2}
\end{figure*}

\begin{table}[htp!]
\scriptsize
\centering
\setlength{\tabcolsep}{2pt}
\begin{tabular}{lcccccccc}
\toprule
\textbf{Method} & \multicolumn{2}{c}{\textbf{Classroom}} & \multicolumn{2}{c}{\textbf{Dampened}} & \multicolumn{2}{c}{\textbf{Hallway}} & \multicolumn{2}{c}{\textbf{Complex}} \\
\cmidrule(r){2-3} \cmidrule(r){4-5} \cmidrule(r){6-7} \cmidrule(r){8-9}
& \textbf{MAG} & \textbf{ENV} & \textbf{MAG} & \textbf{ENV} & \textbf{MAG} & \textbf{ENV} & \textbf{MAG} & \textbf{ENV} \\
\midrule
Random Across & 1.98 & 4.45 & 3.571 & 7.573 & 3.415 & 7.571 & 1.762 & 6.791 \\
Random Same   & 0.710 & 2.182  & 0.213 & 1.635 & 1.104 & 6.582 & 0.685 & 2.831 \\
Linear Interp & 0.725 & 1.890  & 0.110 & 0.908 & 1.082 & 5.566 & 0.637 & 2.370 \\
Nearest Neigh & 0.600 & 2.003 & 0.108 & 0.916 & 0.793 & 5.589 & 0.542 & 2.498 \\
Diff-RIR (K=12) & 0.486 & 1.826 & \textbf{0.085} & \textbf{0.883} & 0.724 & \textbf{5.173} & \textbf{0.442} & 2.197 \\
xRIR (K=8) & \textbf{0.456} & \textbf{1.824}  & 0.093 & 0.892 & \textbf{0.718} & 5.320  & 0.466 & \textbf{2.142} \\
\bottomrule
\end{tabular}
\caption{\textbf{Sim-to-Real Transfer Results} using MAG and ENV metrics on HearingAnythingAnywhere dataset.}
\label{tab:sim_to_real_other}
\end{table}

\begin{table}[h!]
\centering
\begin{tabular}{lccc}
\toprule
\textbf{Method}                  & \textbf{EDT} & \textbf{C50} & \textbf{T60} \\
\midrule
\ourapproach w.o Reference RIRs          & 0.166       & 3.925            & 32.69            \\
\ourapproach w.o Direct Path Module      & 0.061           & 1.596           &  12.33         \\
\ourapproach w.o Reflection Module       & 0.059           & 1.498            &  11.93 \\
\ourapproach (full)                     & \textbf{0.055} & \textbf{1.457} & \textbf{10.53}           \\
\bottomrule
\end{tabular}
\caption{Ablation Study: Comparison of different ablated \ourapproach components on \ourdataset with unseen splits. We report EDT (s), C50 (dB), and T60 (\%) error metrics}
\label{tab:ablate_xrir}
\end{table}

\section{Additional Experiments and Ablation Studies}
\label{sec:ablation}

\textbf{Additional Comparisons on HearingAnythingAnywhere dataset}: We use MAG from ~\cite{chen2024avcloud} and ENV from ~\cite{lan2024acoustic} to further evaluate waveform similarity. For fair comparison, we evaluate our model on the first 0.435s of RIRs due to model length constraints. As shown in Table \ref{tab:sim_to_real_other}, our method performs similarly to DiffRIR on these metrics but outperforms in perceptual metrics like EDT and C50 (Table 2 main paper), which better reflect perceptual quality of rendered RIRs.

\noindent \textbf{Ablation Studies:}\\
We perform ablation studies on the components of \ourapproach by 
considering the followings:

\begin{itemize}
\item \ourapproach w.o Reference RIRs: This ablated variant is the same as our model's adaptation to single-room RIR prediction task. Since there are no reference RIRs as inputs, the model utilizes the implicit neural function to synthesize the target RIR.

\item \ourapproach w.o Direct Path Module: By removing the Direct Path Module, the model only considers the relationship between sources / receivers and  room geometry information, without computing the direct path features.

\item \ourapproach w.o Reflection Module: By removing the Reflection Module, the model only considers the spatial relationship between sources and receivers without taking local geometry information into account.
\end{itemize}

For the above ablation variants, we perform experiments on \ourdataset under unseen settings. We set the number of reference RIRs $K = 8$ for models that take reference RIRs as inputs. As shown in Table \ref{tab:ablate_xrir}, our full model outperforms all ablated variants across all metrics. It is note-worthy that without providing reference RIR as inputs to the model, the model is not able to synthesize reasonable RIRs by just relying on geometric and positional inputs. Also, removing either Direct Path module or Reflection module will lead to degraded performance across all acoustics metrics, demonstrating the importance of capturing full spatial and geometric information for accurate RIR predictions.

Furthermore, we study the importance of finetuning as well as the pretraining on our simulation dataset. In general, we find that simulation data helps the model capture general acoustic properties, such as geometry and material effects. Finetuning on just 12 real samples allows the model to adapt to specific factors like the source’s directivity, improving EDT and C50 metrics compared to training from scratch or without finetuning, as shown in Table \ref{tab:finetune_importance}.\\

\begin{table}[h!]
\centering
\begin{tabular}{lccc}
\toprule
\textbf{Setting} & \textbf{EDT (s)} & \textbf{C50 (dB)} & \textbf{T60 (\%)} \\
\midrule
Scratch    & 0.322 & 4.322  & 7.381 \\
Pretrained & 0.204 & 3.427  & \textbf{4.685} \\
Finetuned  & \textbf{0.092} & \textbf{1.614} & 6.020 \\
\bottomrule
\end{tabular}
\vspace{-2mm}
\caption{Importance of finetuning and pretraining on HearingAnythingAnywhere dataset (classroom).}
\label{tab:finetune_importance}
\end{table}

In addition, to study the impact of scale of \ourdataset on the performance of model, we retrain our model on different number of rooms using xRIR (8-shot) in the unseen setting, while keeping the test split the same. As shown in Table \ref{tab:scale_data}, performance improves with more data but diminishes as room count increases.

\begin{table}[h!]
\centering
\begin{tabular}{lccc}
\toprule
\textbf{Rooms} & \textbf{EDT (s)} & \textbf{C50 (dB)} & \textbf{T60 (\%)} \\
\midrule
65  & 0.088 & 1.813  & 13.79 \\
130 & 0.062 & 1.578  & 11.46 \\
260 & \textbf{0.055} & \textbf{1.457} & \textbf{10.53} \\
\bottomrule
\end{tabular}
\caption{Impact of data scale on model performance.}
\label{tab:scale_data}
\end{table}

\section{Additional Qualitative Samples}
\label{sec:qual_results_suppl}
We provide additional qualitative results of comparisons between our model \ourapproach and the baseline methods on the predicted RIRs on both the simulation dataset and real dataset.

\vspace{0.05in}

As shown in Figure \ref{fig:add_qual_rir_1} and \ref{fig:add_qual_rir_2}, we visualize predicted RIR waveforms versus ground truth RIRs on three simulation environments (apartment, restaurant and bathroom) in \ourdataset as well as two real environments (hallway and classroom) in the Hearing-Anything-Anywhere dataset. At same location in these environments, RIRs predicted by \ourapproach align more closely with the ground truth RIRs, demonstrating the effectiveness of \ourapproach in RIR predictions under both simulated and real settings.

\end{document}